\documentclass[runningheads]{llncs}
\usepackage[T1]{fontenc}
\usepackage{graphicx,verbatim}
\usepackage{subcaption}
\usepackage{amsmath}
\usepackage{amsfonts}
\usepackage{pifont} 
\usepackage{hyperref}
\usepackage{amssymb}
\usepackage{booktabs}
\usepackage{multirow}
\usepackage{amsmath}
\usepackage{float}
\usepackage{capt-of}
\usepackage{adjustbox}
\usepackage[table]{xcolor}

\begin{document}
\title{Attention-Based Prototype Calibration for Multi-Rater Few-Shot Medical Image Segmentation}
%

\author{Truong Vu \and Minh Khoi Ho \and Yutong Xie\thanks{Corresponding author: yutong.xie@mbzuai.ac.ae}}
\authorrunning{Vu et al.}
\institute{Mohamed bin Zayed University of Artificial Intelligence \\
\email{\{truong.vu, khoi.ho, yutong.xie\}@mbzuai.ac.ae}}
  
\maketitle              
\begin{abstract}
Few-shot medical image segmentation methods typically assume a single ground-truth annotation, overlooking systematic variability across expert raters commonly observed in clinical datasets. We propose an attention-based prototype calibration framework for few-shot multi-rater segmentation that models rater-specific deviations from a consensus representation in prototype space. A lightweight yet principled attention operator directly refines rater prototypes without modifying the backbone feature extractor, making the approach fully compatible with existing prototype-based few-shot segmentation methods. This design preserves semantic consistency while enabling personalized segmentation outputs with minimal computational overhead. Experiments on multi-rater medical imaging datasets demonstrate consistent improvements over baseline prototype approaches, highlighting the effectiveness of structured prototype calibration for modeling annotation variability.


\end{abstract}

\section{Introduction}

Medical image segmentation underpins many clinical applications, including tumor assessment and organ volumetry~\cite{banerjee2025novel,liu2025daunet,xia2025comprehensive}. Although deep neural networks achieve strong performance~\cite{azad2024medical,conze2023current,kumar2025advancements}, they rely on densely annotated data, which is expensive and time-consuming to obtain. Few-shot medical image segmentation (FSMIS) mitigates this issue by learning transferable representations that adapt to unseen classes from only a few labeled support images. Prototype-based approaches~\cite{gmrdnet,Ding_2023_WACV,hansen2022anomaly,miningproto,kim2025tied,lin2023few,shen2023q,tang2025few,zhu2023few}, such as PANet~\cite{wang2019panet} and SSL-ALPNet~\cite{sslalpnet}, compute class prototypes from support masks and perform dense feature matching for query segmentation, demonstrating strong generalization under limited supervision.

However, most FSMIS methods assume a single reliable annotation per image. In realistic clinical practice, images are often labeled by multiple experts whose delineations differ systematically due to boundary tolerance and clinical interpretation. Such inter-rater variability, observed in datasets such as CURVAS~\cite{curvas}, LIDC-IDRI~\cite{Armato2011LIDC}, and QUBIQ~\cite{li2024qubiq}, reflects structured stylistic differences rather than random noise. While multi-rater segmentation has been studied in fully supervised settings through consensus estimation~\cite{Warfield2004SimultaneousTA}, uncertainty modeling~\cite{baumgartner2019phiseg,kohl2018probabilistic}, and personalized prediction~\cite{Ji_2021_MRNet,liao2024modeling,Wu_2024_CVPR}, these approaches assume abundant annotations and are not tailored to few-shot learning.

This reveals a key gap: in practice, supervision is both scarce and heterogeneous. To address this, we introduce \emph{few-shot multi-rater segmentation}, where each support image contains multiple annotations and the model produces $R$ personalized predictions per query, one corresponding to each rater style observed in the support set, thereby avoiding consensus collapse and explicitly modeling multiple plausible segmentation modes.

Collapsing multiple annotations into a consensus discards rater-specific information, whereas training raters independently can not exploit shared semantic structure across annotators nor explicitly model the structured differences between rater styles. Hence, we propose a prototype-centric personalization framework for this setting. An attention-based prototype calibration module models structured deviations between rater-specific and consensus prototypes directly in prototype space. A calibration loss regularizes prototype updates to stabilize personalization under limited support data. To enable multi-style supervision from pseudo labels, we further introduce pseudo-style generation, synthesizing multiple rater-style variants from superpixel-based pseudo labels~\cite{superpixel}. Finally, a two-stage training strategy improves optimization stability.

We evaluate our method on two multi-rater benchmarks across modalities: CURVAS (Abdominal CT) and QUBIQ brain-growth (Brain MRI). Using per-rater Dice as the metric, our approach consistently outperforms FSMIS and multi-rater baselines while maintaining robustness in few-shot scenarios. 

Our contributions are as follows: \\
\hspace*{1.5em}- We formalize the novel problem of \emph{few-shot multi-rater segmentation}, bridging realistic multi-annotator supervision with few-shot learning. \\
\hspace*{1.5em}- We propose a prototype-based personalization framework incorporating attention-based prototype calibration, calibration regularization, pseudo-style generation, and two-stage training for solving few-shot multi-rater segmentation by modeling structured inter-rater variability in prototype space. \\
\hspace*{1.5em}- We demonstrate consistent empirical improvements in per-rater segmentation accuracy across multiple multi-rater medical benchmarks.

\section{Method}

\subsection{Problem Formulation}

Few-shot medical image segmentation (FSMIS) aims to learn a model on a training set $\mathcal{D}_{tr}$ that can segment \emph{unseen} semantic classes in a test set $\mathcal{D}_{te}$ given only a few labeled examples of those unseen classes, \emph{without retraining}. The dataset is split into $\mathcal{D}_{tr}$ and $\mathcal{D}_{te}$ with disjoint class sets $\mathcal{C}_{tr}$ and $\mathcal{C}_{te}$, where $\mathcal{C}_{tr}\cap\mathcal{C}_{te}=\varnothing$. Following the standard meta-learning protocol \cite{sslalpnet}, $\mathcal{D}_{tr}$ and $\mathcal{D}_{te}$ are sampled into episodes $\{(\mathcal{S}_i,\mathcal{Q}_i)\}$ for training and evaluation, respectively. Each episode defines an $N$-way $K$-shot segmentation task with episode classes $\{c_j\}_{j=1}^{N}$ (typically $N{=}1$ in medical FSS): the support set is $\mathcal{S}_i=\{(x_s^l,y_s^l(c_j))\}_{j=1,l=1}^{N, K}$ and the query set $\mathcal{Q}_i$ contains $V$ image-mask pairs from the same class as the support set, where $y_s^l(c_j)\in\{0,1\}^{H\times W}$ is the binary mask for class $c_j$ (background $c_0$ is implicit). A conventional single-label FSMIS model learns to predict one mask $\hat{y}_q^v(c_j)$ for each query image $x_q^v$ ($v=1,\dots,V$) conditioned on $\mathcal{S}_i$. After a series of episodes, we obtain the final segmentation model, which is evaluated on unseen $\mathcal{C}_{te}$ in the same $N$-way $K$-shot segmentation manner.

In realistic multi-rater datasets, each image may be annotated by multiple experts with systematic stylistic differences. We therefore extend to \emph{few-shot multi-rater segmentation}, where each support image has $R$ masks for each class $c_j$ $\{y_s^{l,(r)}(c_j)\}_{r=1}^{R}$, i.e., $\mathcal{S}_i=\{(x_s^l,\{y_s^{l,(r)}(c_j)\}_{r=1}^{R})\}_{j=1,l=1}^{N, K}$. The class split and episodic protocol remain unchanged, but the goal differs: given $x_q^v$, the model must output $R$ personalized predictions $\{\hat{y}_q^{v,(r)}(c_j)\}_{r=1}^{R}$, one per rater style observed in the support set, rather than a single consensus segmentation. Following previous works~\cite{lin2023few,sslalpnet,tang2025few,zhu2023few}, we consider $N=K=1$.

\subsection{Prototype-Based Few-Shot Segmentation}

In this section, we present a simple baseline approach to the \textit{few-shot multi-rater segmentation} problem. This approach is a direct extension of existing prototype-based FSMIS methods~\cite{lin2023few,sslalpnet,tang2025few,zhu2023few}, simply processing each rater independently.

Let $f_\theta(\cdot)$ denote a shared encoder producing feature maps $F_s = f_\theta(x_s), F_q = f_\theta(x_q)$ with $x_s$ and $x_q$ being a support image and a query image from the same episode. Modern few-shot segmentation methods construct prototypes through various modules, including masked pooling, grid-based local prototypes, or multi-scale refinement~\cite{gmrdnet,Ding_2023_WACV,hansen2022anomaly,miningproto,kim2025tied,lin2023few,shen2023q,tang2025few,zhu2023few}.  We denote this prototype extraction function abstractly as $\{p_c^{(r)}\} = \Phi(F_s, y_s^{(r)})$, where $c \in \{\text{fg}, \text{bg}\}$ and $\Phi(\cdot, \cdot)$ may produce one or multiple prototypes per class. The shape of $p_c^{(r)}$ is $N \times D$ with $N$ being the number of prototypes of class $c$ and $D$ being the dimension.

A pixel-wise cosine similarity: $\ell_{q,c}^{(r)}(i) = \text{sim}\big(F_q(i), p_c^{(r)}\big)$ is used to compute query logits, where $i$ is the spatial location on the feature map. The foreground logit $\ell_{q,\text{fg}}^{(r)}(i)$ and background logit $\ell_{q,\text{bg}}^{(r)}(i)$ are stacked to form $\ell_q^{(r)}(i)$.
Class probabilities are obtained via softmax: $\pi_q^{(r)}(i) = \text{softmax}(\ell_q^{(r)}(i))$, where $\pi_q^{(r)}\in\mathbb{R}^{H\times W\times C}$ and $\pi_{q,c}^{(r)}$ denotes its $c$-channel ($c\in \{\text{fg,bg}\}$).
The final predicted mask for rater $r$ is defined as $\hat{y}_q^{(r)}(i) = \underset{c\in \{\text{fg,bg}\}}{\arg\max}\ \pi_{q,c}^{(r)}(i)$. 


\begin{figure}[t!]
\centering

\begin{subfigure}[t]{0.7\textwidth}
    \centering
    \includegraphics[width=\linewidth]{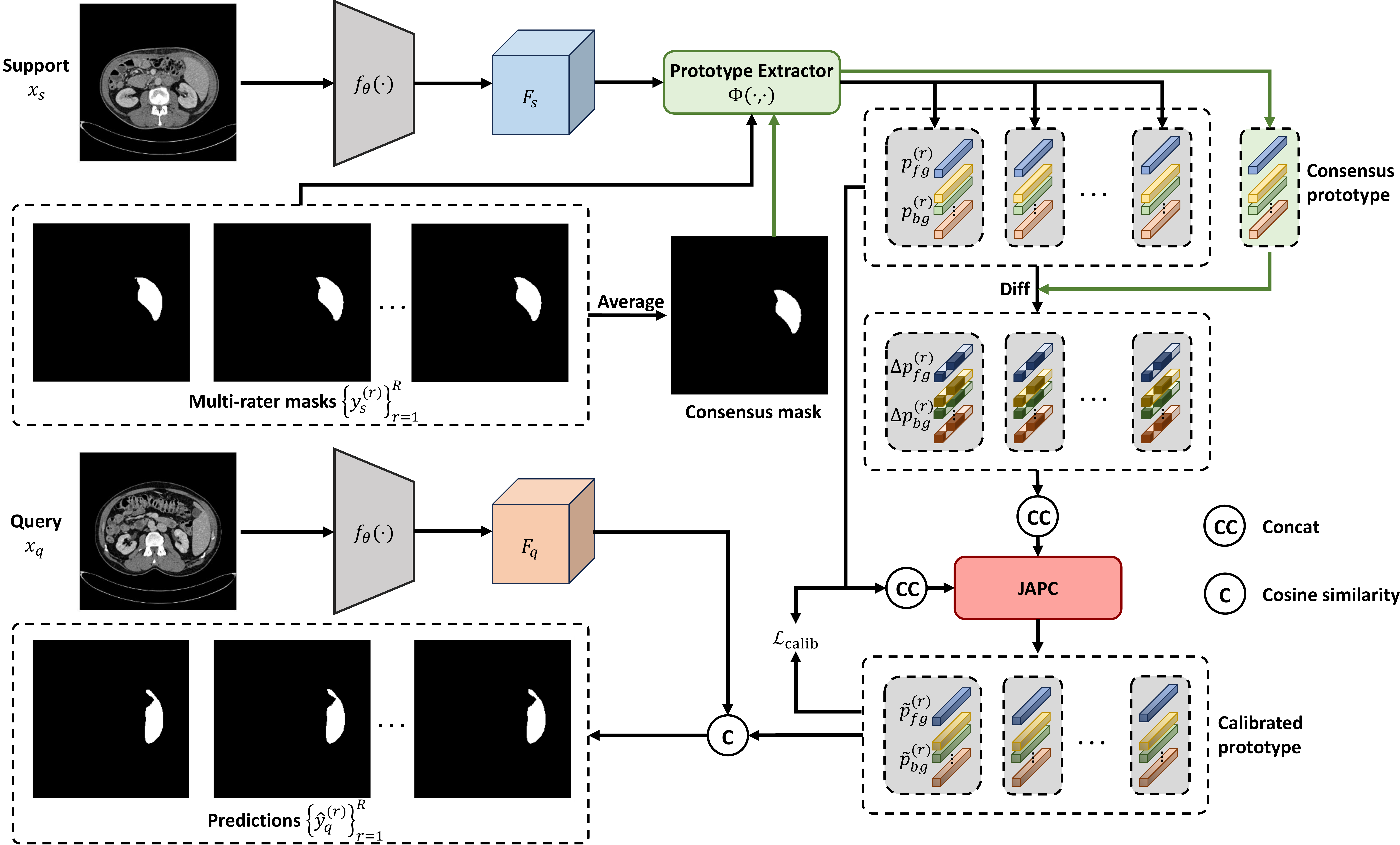}
    \caption{}
    \label{fig:overview}
\end{subfigure}
\hfill
\begin{subfigure}[t]{0.28\textwidth}
    \centering
    \includegraphics[width=\linewidth]{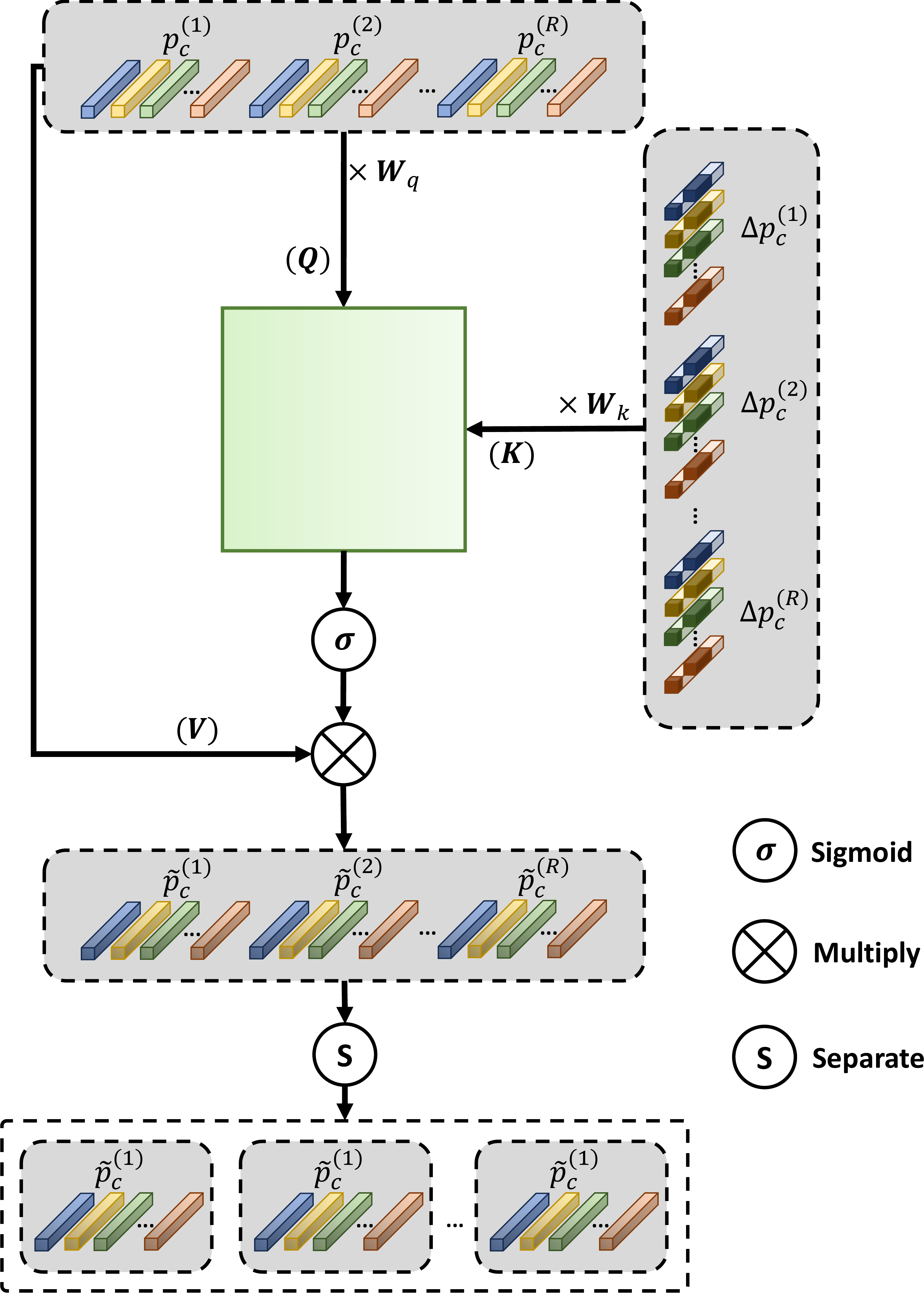}
    \caption{}
    \label{fig:japc}
\end{subfigure}


\begin{subfigure}[t]{\textwidth}
    \centering
    \includegraphics[width=\linewidth]{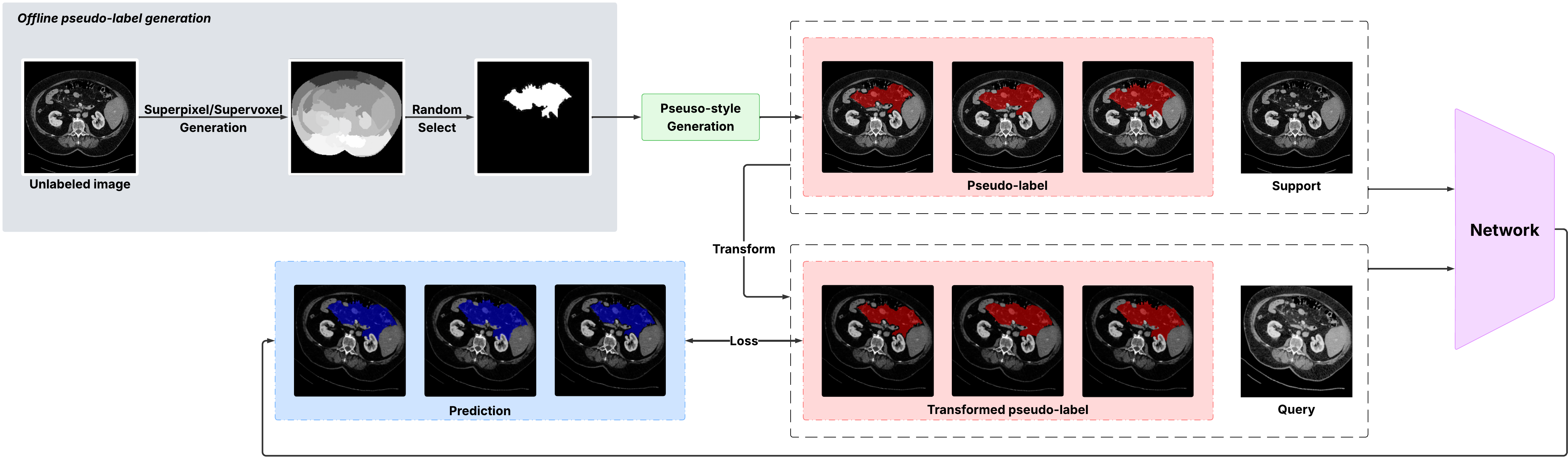}
    \caption{}
    \label{fig:ssl}
\end{subfigure}

\caption{Overview of the proposed framework. (a) Few-shot multi-rater segmentation pipeline. Rater-specific and consensus prototypes are extracted from support features, and calibrated by the \textbf{J}oint \textbf{A}ttention-based \textbf{P}rototype \textbf{C}alibration (\textbf{JAPC}) module to generate $R$ personalized predictions via cosine similarity with query features. (b) Internal structure of \textbf{JAPC}, modeling structured interactions between rater prototypes and their deviations. (c) Pseudo-style generation. Pseudo labels derived from superpixel/supervoxel segmentation are transformed to synthesize multiple rater-style variants, which are used as support labels and query supervision during training. The ``Network'' block in (c) corresponds to the full pipeline shown in (a).}
\label{fig:combined}
\end{figure}

\subsection{JAPC: Joint Attention-based Prototype Calibration}

As mentioned previously, a straightforward baseline to adapt a conventional single-label few-shot model to our novel setting is to treat each rater independently: for each episode, one can process a single rater mask at a time and repeat this for all $R$ raters. 
While simple, this strategy ignores structured relationships among raters. 
In particular, it fails to exploit shared semantic information and does not explicitly model systematic boundary deviations across raters, leading to inefficient learning of stylistic differences, especially in ambiguous regions.

To explicitly model structured inter-rater relationships, we introduce \textbf{JAPC}: a \textbf{J}oint \textbf{A}ttention-based \textbf{P}rototype \textbf{C}alibration module. This module aims to calibrate each rater's prototypes, highlighting the differences among raters. The module is lightweight and can be plugged into any existing prototype-based FSMIS method. An overview of the overall framework and the JAPC module is illustrated in Fig.~\ref{fig:overview} and Fig.~\ref{fig:japc}, respectively.

First, we compute a consensus mask $y_s^{(0)} = \frac{1}{R} \sum_{r=1}^{R} y_s^{(r)}$, and derive a consensus prototype $p_c^{(0)} = \Phi(F_s, y_s^{(0)})$ where $c \in \{\text{fg}, \text{bg}\}$. For each rater and class, we define deviation $\Delta_c^{(r)} = p_c^{(r)} - p_c^{(0)}$. All rater prototypes are concatenated: $P_c = \text{concat}([p_c^{(1)}, \dots, p_c^{(R)}])$. The shape of $P_c$ is $N_{\text{total}} \times D$ with $N_{\text{total}}$ being the total number of prototype vectors of all raters for class $c$ and $D$ being the prototype dimension.

We apply a shared self-attention module for both background prototypes $P_{\text{bg}}$ and foreground prototype $P_{\text{fg}}$:
$
\mathbf{Q} = P_c W_q, \quad
\mathbf{K} = \Delta_c W_k, \quad
\mathbf{V} = P_c ,
$
\[
\mathbf{A} = \sigma\!\left(\mathbf{Q}\mathbf{K}^\top / \sqrt{D}\right), \quad \tilde{P}_c = \mathbf{A} \mathbf{V}.
\]

This joint formulation allows each rater prototype to attend to the deviations of all other raters, enabling structured modeling of shared and distinctive boundary tendencies. 
Calibrated prototypes of each rater $\tilde{p}_c^{(r)}$ are separated from $\tilde{P}_c$ and then used for query prediction:
$
\ell_{q,c}^{(r)}(i) = \text{sim}(F_q(i), \tilde{p}_c^{(r)}).
$

To prevent excessive deformation from semantic anchors, we introduce calibration loss: 
$
\mathcal{L}_{\text{calib}} =
\frac{1}{R} \sum_{r=1}^{R}
\sum_{c}
\| p_c^{(r)} - \tilde{p}_c^{(r)} \|_2^2.
$

\subsection{Pseudo-Style Generation from Pseudo Labels}



Prior work shows that superpixel-derived pseudo labels yield more diverse and generalizable representations than raw annotations~\cite{sslalpnet}; we therefore adopt pseudo labels as primary supervision. However, they are style-agnostic and lack rater-specific variations. To enable personalized modeling while retaining pseudo supervision, we introduce pseudo-style generation. Given a pseudo label $y^{\text{pseudo}}$, we synthesize $R$ style-conditioned variants (with $R$ raters) as $\tilde{y}^{(r)}=\mathcal{T}_r(y^{\text{pseudo}})$, where $\mathcal{T}_r$ denotes a random structured boundary transformation (e.g., dilation/erosion). The transformations vary across episodes to avoid overfitting and simulate plausible inter-rater deviations while preserving semantics. They are applied only during training for support labels and query supervision (see Fig.~\ref{fig:ssl}).


\subsection{Two-Stage Training Strategy}

Directly training the calibration module jointly with a randomly initialized encoder leads to unstable prototype dynamics under few-shot supervision. 
We therefore adopt a two-stage training strategy within a total budget of $T$ iterations: the base prototype model is trained without JAPC and pseudo-style for the first $T/2$ iterations, after which they are enabled and training continues for the remaining $T/2$ iterations.

This strategy improves stability because intermediate encoder features already encode meaningful semantic structure, reducing early prototype noise. 
It also prevents catastrophic drift of prototypes during calibration, enabling effective modeling of subtle stylistic differences at object boundaries.

\subsection{Overall Objective}


The rater-specific segmentation loss is defined as pixel-wise cross-entropy:
$\mathcal{L}_{\text{seg}}^{(r)}
=
\frac{1}{|\Omega|}
\sum_{i \in \Omega}
\text{CE}\big(
\text{softmax}(\ell_q^{(r)}(i)),
y_q^{(r)}(i)
\big)$,
 where $\Omega$ denotes spatial locations.
The overall segmentation loss is averaged across raters: 
$\mathcal{L}_{\text{seg}}
=
\frac{1}{R}
\sum_{r=1}^{R}
\mathcal{L}_{\text{seg}}^{(r)}.
$

Following previous works \cite{sslalpnet,tang2025few,wang2019panet}, we also employ the alignment loss, where the query images act as the support set to predict labels of the support images. Please note that the prototypes in this reverse learning process are also calibrated as in the forward process:
\[\mathcal{L}_{\text{align}}
=
\frac{1}{R}
\sum_{r=1}^{R}
\mathcal{L}_{\text{align}}^{(r)}
=
\frac{1}{R|\Omega|}
\sum_{r=1}^{R}
\sum_{i \in \Omega}
\text{CE}\big(
\text{softmax}(\ell_s^{(r)}(i)),
y_s^{(r)}(i)
\big).
\]

The final objective is $
\mathcal{L} =
\mathcal{L}_{\text{seg}}
+ \mathcal{L}_{\text{align}}
+  \mathcal{L}_{\text{calib}}.
$

\section{Experiments}




\subsection{Datasets}

We evaluate on two multi-rater medical segmentation datasets spanning different modalities and numbers of raters: \textbf{(i)} \textbf{CURVAS}~\cite{curvas} (abdominal CT), termed Abd-CT, contains three organs (kidney (\textit{K}), pancreas (\textit{P}), liver (\textit{L})), each annotated by three experts. We follow the official split with 20 training and 65 test scans. \textbf{(ii)} \textbf{QUBIQ Brain-Growth}~\cite{li2024qubiq} (Brain-MRI) includes one class (brain-growth  (\textit{B})) annotated by seven raters. As the official test set is unavailable, we use the validation set for evaluation, resulting in 34 training and 5 test scans.
\subsection{Evaluation Metrics}

Following previous works\cite{sslalpnet,tang2025few,Wu_2024_CVPR}, we evaluate segmentation performance using the Dice score (\%). Suppose $\text{Dice}_{R_i}^c$ is the Dice score (\%) of rater $i$ with respect to class $c$. For each dataset, we report $\text{Dice}_{m}^c$ (the average Dice across raters for class $c$), $\text{Dice}_{m}^{R_i}$ (the average Dice across classes for rater $R_i$), and $\text{Dice}_{m}=\frac{1}{C}\sum_c \text{Dice}_{m}^c$ (the average Dice across all classes and raters).

\subsection{Few-shot Settings}



We follow two common few-shot evaluation protocols~\cite{sslalpnet,roy2020squeeze,tang2025few}. In \textbf{Setting 1 (standard)}, the model is trained and tested on all classes without explicit class partitioning; test classes may appear as background during training but are not supervised as foreground. In \textbf{Setting 2 (strict unseen-class)}, test classes are entirely excluded from training, and any image containing test-class pixels is removed to ensure true class disjointness. Label partitioning differs accordingly. In Setting 1, pseudo-label training requires no partitioning. In Abd-CT under Setting 2, pancreas and liver often co-occur in 2D slices, so they are grouped (upper abdomen); when testing one group, all its classes are excluded from training. Since Brain-MRI contains only one class, it is evaluated only under Setting 1.

\begin{table}[t!]
\centering
\small
\caption{Setting 1 results. We report per-class mean Dice on Abd-CT and Brain-MRI. Voxel-based (supervoxel) methods are not applicable to Brain-MRI as images are single-sliced. Best and second best values are \textbf{bold} and \underline{underlined}.}
\label{tab:curvas_qubiq_meanonly}
\resizebox{0.7\textwidth}{!}{%
\begin{tabular}{@{}ll cccc c@{}}
\toprule
\multirow{2}{*}{Settings} & \multirow{2}{*}{Method}
& \multicolumn{4}{c}{Abd-CT}
& \multicolumn{1}{c}{Brain-MRI} \\
\cmidrule(lr){3-6}\cmidrule(lr){7-7}
& &
$\mathrm{Dice}^{K}_{m}$ &
$\mathrm{Dice}^{P}_{m}$ &
$\mathrm{Dice}^{L}_{m}$ &
$\mathrm{Dice}_{m}$ &
$\mathrm{Dice}^{B}_{m}$ \\
\midrule

\rowcolor{gray!10}\textit{Full-sup}
& Backbone
& 85.02 & 57.46 & 91.40 & 77.96
& 87.42 \\
\midrule\midrule

\textit{Multi-rater}
& MRNet~\cite{Ji_2021_MRNet}
& 56.19 & 15.20 & 68.29 & 46.56
& 62.35 \\
& D-Persona~\cite{Wu_2024_CVPR}
& 64.72 & 26.67 & 67.16 & 53.05
& 57.20 \\
\midrule

\multirow{2}{*}{\shortstack[l]{\textit{Few-shot}\\\textit{supervoxel}}}
& Q-Net~\cite{shen2023q}
& 61.06 & \underline{39.56} & 57.44 & 52.69
& \textit{--} \\
& RPT~\cite{zhu2023few}
& 63.82 & \textbf{41.87} & 71.66 & 59.12
& \textit{--} \\
\midrule

\multirow{4}{*}{\shortstack[l]{\textit{Few-shot}\\\textit{superpixel}}}
& SSL-ALPNet~\cite{sslalpnet}
& \underline{69.33} & 37.68 & 77.42 & 61.48
& 61.37 \\
& SSL-ALPNet~\cite{sslalpnet} + Ours
& \textbf{71.04} & 39.06 & 78.26 & \textbf{62.79}
& \underline{65.78} \\
\cmidrule{2-7}
& DSPNet~\cite{tang2025few}
& 68.03 & 37.96 & \underline{78.58} & 61.52
& 54.02 \\
& DSPNet~\cite{tang2025few} + Ours
& 68.63 & 39.09 & \textbf{79.03} & \underline{62.25}
& \textbf{66.62} \\
\bottomrule
\end{tabular}%
}
\end{table}



\subsection{Implementation Details}

We compare with representative FSMIS methods, including SSL-ALPNet~\cite{sslalpnet} and DSPNet~\cite{tang2025few} (superpixel-based), Q-Net~\cite{shen2023q} and RPT~\cite{zhu2023few} (supervoxel-based), as well as multi-rater methods MRNet~\cite{Ji_2021_MRNet} and D-Persona~\cite{Wu_2024_CVPR} trained on the support set. All few-shot methods use identical support–query splits and share the same DeepLabv3-ResNet101~\cite{chen2017rethinking} backbone. A fully supervised model is included as an upper bound. We adopt default hyperparameters for all benchmarks. On Abd-CT, FSMIS models are trained for 50k iterations, and ours initializes from the 25k checkpoint following the two-stage schedule. On Brain-MRI, FSMIS models are trained for 10k iterations, and ours starts from 5k. Multi-rater models and fully supervised baseline are trained for 50 epochs each.

\subsection{Results}

\begin{figure}[t!]
    \centering
    \begin{minipage}[t]{0.54\columnwidth}
        \centering
        \vspace{0pt}
        \includegraphics[width=\linewidth]{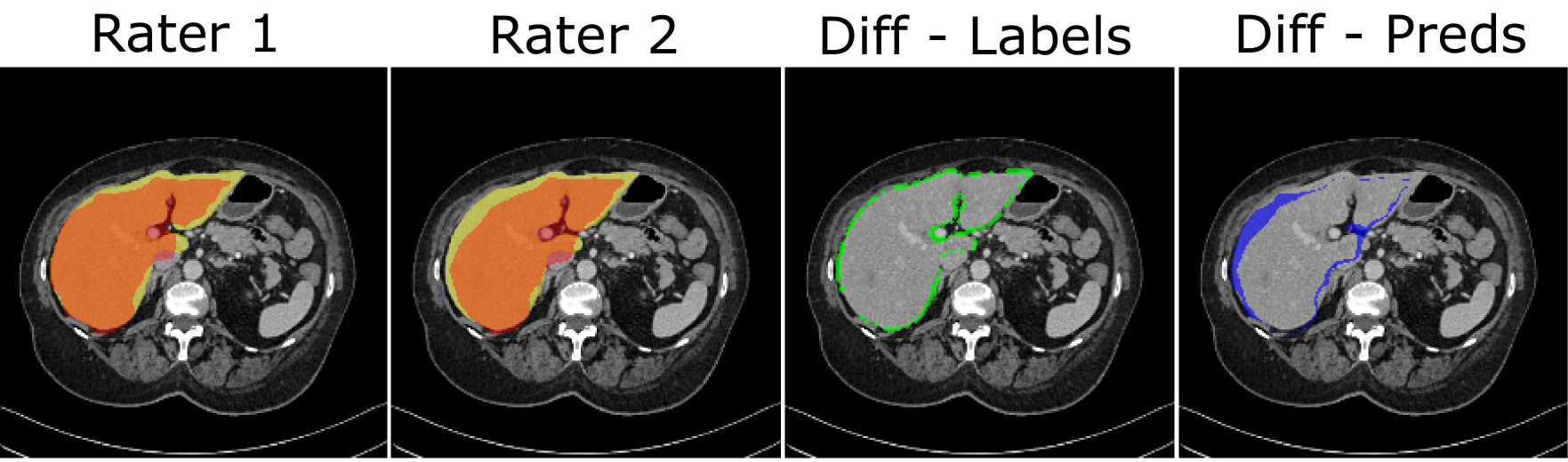}
    \end{minipage}\hfill
    \begin{minipage}[t]{0.44\columnwidth}
        \centering
        \vspace{0pt}
        \scriptsize
        \setlength{\tabcolsep}{3pt}
        \renewcommand{\arraystretch}{1.1}

        \begin{adjustbox}{max width=\linewidth}
        \begin{tabular}{@{}lcccc@{}}
            \toprule
            Method & $\mathrm{Dice}^{R_1}_{m}$ & $\mathrm{Dice}^{R_2}_{m}$ & $\mathrm{Dice}^{R_3}_{m}$ & $\mathrm{Dice}_{m}$ \\
            \midrule
            Q-Net~\cite{shen2023q}
            & 54.73 & 54.80 & 51.85 & 53.79 \\
            RPT~\cite{zhu2023few}
            & 51.59 & 51.70 & 49.79 & 51.03 \\
            \midrule
            SSL-ALPNet~\cite{sslalpnet}
            & 57.01 & 58.08 & 54.65 & 56.58 \\
            \textbf{+ Ours}
            & \underline{59.23} & \underline{59.95} & \underline{56.88} & \underline{58.69} \\
            \midrule
            DSPNet~\cite{tang2025few}
            & 58.62 & 59.18 & 56.54 & 58.11 \\
            \textbf{+ Ours}
            & \textbf{60.23} & \textbf{60.82} & \textbf{57.82} & \textbf{59.62} \\
            \bottomrule
        \end{tabular}
        \end{adjustbox}
    \end{minipage}

    \caption{\textit{(left)} Qualitative example with ground truth (yellow) and predictions (red) and their rater-wise pixel difference (green \& blue respectively). \textit{(right)} Per-rater macro-averaged Dice over Kidney/Pancreas/Liver on Abd-CT Setting~2.}
    \label{fig:qualitative}
\end{figure}

\paragraph{Few-shot performance.} Across both settings, our method consistently improves strong few-shot baselines on Abd-CT and Brain-MRI (Tables~\ref{tab:curvas_qubiq_meanonly} and~\ref{tab:curvas_s2_meanonly}). On Abd-CT, it yields consistent gains over SSL-ALPNet and DSPNet in both standard and strict unseen-class settings. On Brain-MRI, improvements are more pronounced, particularly over DSPNet, demonstrating the effectiveness of personalized prototype calibration under multi-rater supervision.

\paragraph{Rater-wise robustness.} Improvements are consistent across annotators (Fig.~\ref{fig:qualitative}), indicating stable personalization rather than gains limited to specific raters. Qualitative results further show that our predictions better capture shared structures while adapting to boundary variations among raters.

\paragraph{Ablation study.} Table~\ref{tab:ablation_s2_meanonly} verifies the contribution of each component on Abd-CT Setting~2. Pseudo-style supervision provides a modest improvement, attention and two-stage training further enhance performance, and calibration loss significantly strengthens prototype regularization. Combining all modules yields the best overall result, confirming their complementary effects.

\paragraph{Limitations.} A limitation is that performance on pancreas may still lag behind supervoxel-based methods, which benefit from stronger region-level constraints; although pseudo-style supervision reduces fragmentation, it does not always eliminate this gap.


\begin{table*}[t!]
\centering

\begin{minipage}[t]{0.49\textwidth}
    \centering
    \vspace{0pt}
    \captionof{table}{Abd-CT Setting 2 with mean Dice for each organ.}
    \label{tab:curvas_s2_meanonly}

    \scriptsize
    \setlength{\tabcolsep}{4pt}
    \renewcommand{\arraystretch}{1.15}

    \begin{adjustbox}{max width=\linewidth}
    \begin{tabular}{@{}lcccc@{}}
        \toprule
        Method & Dice$^{K}_{m}$ & Dice$^{P}_{m}$ & Dice$^{L}_{m}$ & Dice$_m$ \\
        \midrule
        Q-Net~\cite{shen2023q}             & 47.86 & 36.33 & 77.19 & 53.79 \\
        RPT~\cite{zhu2023few}               & 50.29 & \textbf{39.88} & 62.91 & 51.03 \\
        SSL-ALPNet~\cite{sslalpnet}        & 58.50 & 38.23 & 73.02 & 56.58 \\
        SSL-ALPNet~\cite{sslalpnet}  + Ours & 61.49 & \underline{38.24} & 76.33 & \underline{58.69} \\
        DSPNet~\cite{tang2025few}            & \underline{62.24} & 35.32 & \underline{76.77} & 58.11 \\
        DSPNet~\cite{tang2025few} + Ours     & \textbf{64.10} & 36.78 & \textbf{77.98} & \textbf{59.62} \\
        \bottomrule
    \end{tabular}
    \end{adjustbox}
\end{minipage}
\hfill
\begin{minipage}[t]{0.49\textwidth}
    \centering
    \vspace{0pt}
    \captionof{table}{Ablation study on Abd-CT Setting 2 of DSPNet.}
    \label{tab:ablation_s2_meanonly}

    \scriptsize
    \setlength{\tabcolsep}{3pt}
    \renewcommand{\arraystretch}{1.15}

    \begin{adjustbox}{max width=\linewidth}
    \begin{tabular}{@{}cccccccc@{}}
        \toprule
        \multirow{2}{*}{\shortstack[c]{Pseudo\\style}} &
        \multirow{2}{*}{Attn} &
        \multirow{2}{*}{\shortstack[c]{2-\\stage}} &
        \multirow{2}{*}{\shortstack[c]{Calib\\loss}} &
        \multirow{2}{*}{Dice$^{K}_{m}$} &
        \multirow{2}{*}{Dice$^{P}_{m}$} &
        \multirow{2}{*}{Dice$^{L}_{m}$} &
        \multirow{2}{*}{Dice$_m$} \\
        & & & & & & & \\
        \midrule
        $\times$ & $\times$ & $\times$ & $\times$ & 62.24 & 35.32 & 76.77 & 58.11 \\
        $\checkmark$ & $\times$ & $\times$ & $\times$ & 62.43 & 34.21 & \underline{78.32} & 58.32 \\
        $\checkmark$ & $\checkmark$ & $\times$ & $\checkmark$ & \textbf{66.00} & 34.94 & 77.05 & \underline{59.33} \\
        $\checkmark$ & $\checkmark$ & $\checkmark$ & $\times$ & 62.86 & \underline{36.53} & \textbf{78.48} & 59.29 \\
        $\checkmark$ & $\checkmark$ & $\checkmark$ & $\checkmark$ & \underline{64.10} & \textbf{36.78} & 77.98 & \textbf{59.62} \\
        \bottomrule
    \end{tabular}
    \end{adjustbox}
\end{minipage}

\end{table*}

\section{Conclusion}

We introduced the problem of few-shot multi-rater medical image segmentation, bridging the gap between conventional few-shot learning and realistic multi-annotator supervision. To address the challenges of heterogeneous annotations under limited support data, we proposed a prototype-based personalization framework with pseudo-style generation, joint attention-based prototype calibration, calibration regularization, and two-stage training. The proposed method explicitly models structured inter-rater variations while preserving shared semantic representations. Experiments on CURVAS and QUBIQ demonstrate consistent improvements in Dice scores across modalities and anatomical structures. Our results highlight the importance of modeling annotation heterogeneity in few-shot regimes and provide a practical foundation for personalized few-shot medical segmentation.

%
%
%
\newpage
\bibliographystyle{splncs04}
\bibliography{mybibliography}
%




\end{document}